\documentclass[10pt,twocolumn,letterpaper]{article}

\usepackage{times}
\usepackage[T1]{fontenc}
\usepackage[letterpaper,top=0.8in,bottom=0.8in,left=0.75in,right=0.75in,columnsep=0.28in]{geometry}
\usepackage{amsmath,amssymb}
\usepackage{graphicx}
\usepackage{booktabs,multirow}
\usepackage[dvipsnames]{xcolor}
\usepackage{xspace}
\usepackage[numbers,sort&compress]{natbib}
\usepackage[font=small,labelfont=bf,labelsep=period]{caption}
\usepackage[hyphens]{url}
\definecolor{preprintlink}{RGB}{25,75,120}
\usepackage[colorlinks=true,allcolors=preprintlink,breaklinks=true]{hyperref}
\hypersetup{
    pdftitle={NereidVLA: Learning to Understand and Navigate the Underwater World},
    pdfauthor={Chenjie Zhao, Zhiliang Zhu, Quan Chen, Yuquan Xue, Xingxue Shi, Puyue Wang, Zhiqiang Zhao, Kangye Zheng, Boyang Xu, Mingwei Lin, Dejun Li}
}
\makeatletter
\renewcommand\section{\@startsection{section}{1}{\z@}%
    {-2.5ex plus -0.6ex minus -0.2ex}{1.1ex plus 0.2ex}{\normalfont\large\bfseries}}
\renewcommand\subsection{\@startsection{subsection}{2}{\z@}%
    {-2.0ex plus -0.5ex minus -0.2ex}{0.8ex plus 0.2ex}{\normalfont\normalsize\bfseries}}
\renewcommand\paragraph{\@startsection{paragraph}{4}{\z@}%
    {0.8em}{-1em}{\normalfont\normalsize\bfseries}}
\makeatother
\newcommand{\etal}{\emph{et al.}\xspace}

\usepackage{graphicx}
\usepackage{hyperref}

\newcommand{\PreprintAuthors}{%
    Chenjie Zhao\textsuperscript{1,*},
    Zhiliang Zhu\textsuperscript{2,*,\textdagger,\textdaggerdbl},
    Quan Chen\textsuperscript{1},
    Yuquan Xue\textsuperscript{3},\\[0.25em]
    Xingxue Shi\textsuperscript{4},
    Puyue Wang\textsuperscript{5},
    Zhiqiang Zhao\textsuperscript{2},
    Kangye Zheng\textsuperscript{2},\\[0.25em]
    Boyang Xu\textsuperscript{2},
    Mingwei Lin\textsuperscript{1,\textdagger},
    Dejun Li\textsuperscript{1}%
}

\newcommand{\PreprintAffiliations}{%
    \textsuperscript{1}Zhejiang University
    \quad
    \textsuperscript{2}Harbin Engineering University
    \\[0.2em]
    \textsuperscript{3}Nanyang Technological University
    \quad
    \textsuperscript{4}Harbin Institute of Technology
    \\[0.2em]
    \textsuperscript{5}University of Auckland%
}

\newcommand{\PreprintAuthorNotes}{%
    \textsuperscript{*}Equal contribution.\quad
    \textsuperscript{\textdagger}Corresponding authors.\quad
    \textsuperscript{\textdaggerdbl}Project leader.%
}

\newcommand{\UniversityLogoGap}{25pt}

\title{NereidVLA: Learning to Understand and Navigate the Underwater World}
\date{}

\newif\ifpreprintappendix
\preprintappendixtrue

\begin{document}

\twocolumn[{

    \noindent
    \makebox[\textwidth][c]{%
        \raisebox{-0.5\height}{%
            \includegraphics[
                height=48pt,
                trim=64bp 30bp 70bp 39bp,
                clip
            ]{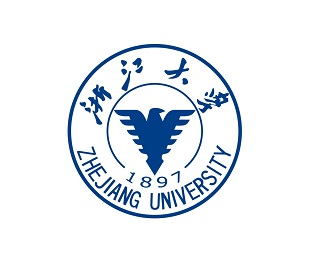}%
        }%
        \hspace{\UniversityLogoGap}%
        \raisebox{-0.5\height}{%
            \includegraphics[
                height=55pt
            ]{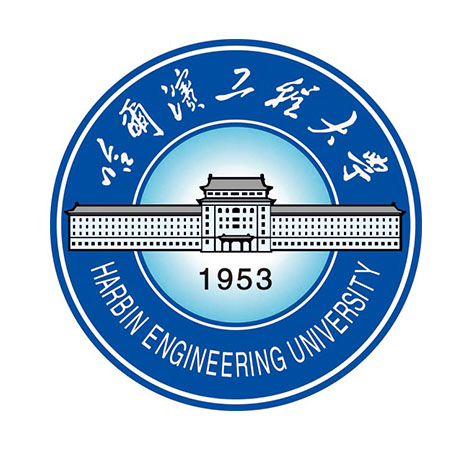}%
        }%
        \hspace{\UniversityLogoGap}%
        \raisebox{-0.5\height}{%
            \includegraphics[
                height=46pt
            ]{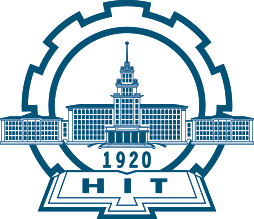}%
        }%
        \hspace{\UniversityLogoGap}%
        \raisebox{-0.5\height}{%
            \includegraphics[
                height=41pt
            ]{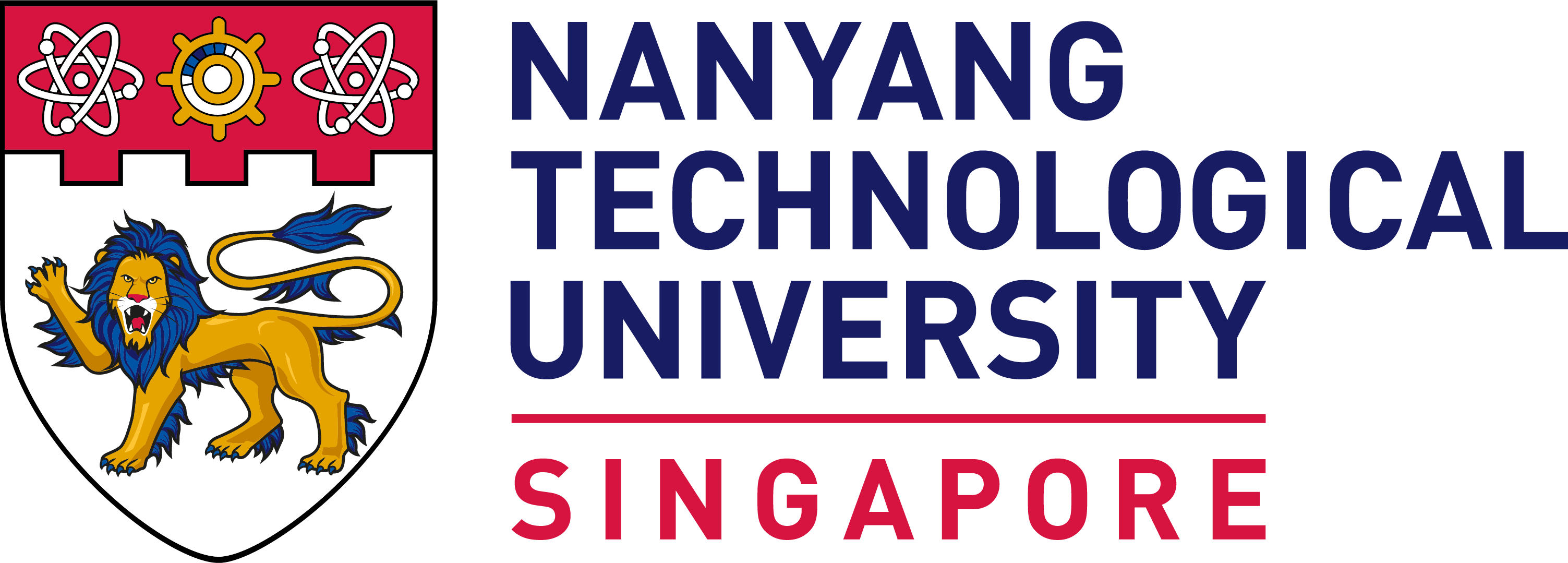}%
        }%
        \hspace{\UniversityLogoGap}%
        \raisebox{-0.5\height}{%
            \includegraphics[
                height=50pt
            ]{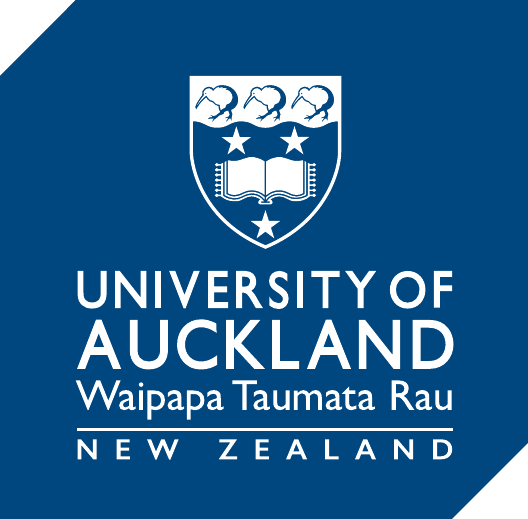}%
        }%
    }\par

    \vspace{-0.6em}

    \begin{center}

        {\color{preprintlink}\rule{\textwidth}{1.6pt}}\par
        \vspace{0.8em}

        {\fontsize{20}{24}\selectfont\bfseries
            \raisebox{-5pt}{%
                \includegraphics[
                    height=25pt
                ]{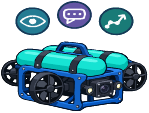}%
            }\hspace{0.3em}%
            \raisebox{-1pt}{%
                \includegraphics[
                    height=19pt
                ]{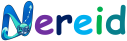}%
            }\hspace{0.04em}%
            VLA: Learning to Understand\par
            and Navigate the Underwater World\par
        }

        \vspace{0.8em}
        {\color{preprintlink}\rule{\textwidth}{0.45pt}}\par
        \vspace{0.8em}

        {\fontsize{10}{14}\selectfont
            \PreprintAuthors\par
        }

        \vspace{0.65em}

        {\small
            \PreprintAffiliations\par
        }

        \vspace{0.4em}

        {\footnotesize
            \PreprintAuthorNotes\par
        }

        \vspace{0.35em}
        {\footnotesize
            \href{mailto:zhao_chenjie@zju.edu.cn}{%
                \texttt{zhao\_chenjie@zju.edu.cn}%
            }
            \quad
            \href{mailto:zhuzhiliang@hrbeu.edu.cn}{%
                \texttt{zhuzhiliang@hrbeu.edu.cn}%
            }\par
        }

        \vspace{0.4em}
        {\small
            \href{https://github.com/SKY-Zhu-91/NereidVLA}{%
                \textcolor{preprintlink}{%
                    \textbf{GitHub:}\;
                    \texttt{github.com/SKY-Zhu-91/NereidVLA}%
                }%
            }\par
        }

    \end{center}

    \vspace{0.3em}

    \begin{minipage}{\textwidth}
        \begin{abstract}
Underwater navigation requires translating visual spatial cues into motion, yet marine vision-language datasets rarely include paired robot trajectories.
We present NereidVLA, a vision-language-action model that transfers marine spatial knowledge to language-conditioned 3D navigation.
We first train Sea2Know on approximately two million marine examples to learn target grounding, spatial relations, and scene structure without paired robot-motion supervision.
We then adapt it using expert trajectories executed under simulated underwater vehicle dynamics.
Given RGB histories and a language instruction, NereidVLA predicts discrete motion tokens that decode into local 3D trajectories for closed-loop tracking.
Evaluation combines an independent six-dimensional underwater navigation understanding benchmark with offline prediction and closed-loop navigation.
Under matched navigation training and execution settings, marine spatial pretraining increases simulation success from 72.4\% to 84.4\%.
Without fine-tuning on the real test environment, NereidVLA achieves 78.75\% success in laboratory-pool robot trials, compared with 47.50\% for InternVLA-N1.
These results show that marine spatial supervision can support downstream motion learning and physical underwater navigation.
\end{abstract}
    \end{minipage}

    \vspace{1.4em}

}]

\section{Introduction}
\label{sec:intro}

Language-guided underwater navigation requires robots to interpret a goal
in the context of three-dimensional scene geometry. Following a pipeline
requires understanding its direction and the obstacles nearby, while cave
navigation depends on recognizing entrances and the geometry of traversable
passages. These decisions involve changes in depth as well as horizontal
motion: recognizing the target is only one part of determining how to reach
it. Turbidity, light scattering, changing illumination, and occlusion further
complicate these spatial judgments~\cite{akkaynak2018underwater}.

\begin{figure*}[!t]
    \centering
    \includegraphics[width=\textwidth]{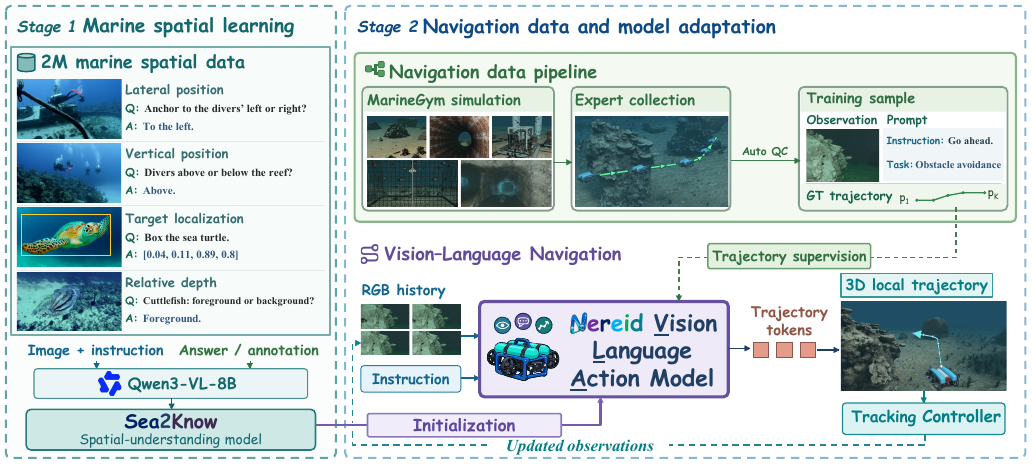}
    \caption{Overview of the two-stage learning framework for NereidVLA.
    Marine spatial learning produces Sea2Know, which initializes the navigation
    model. Expert demonstrations executed under simulated vehicle dynamics
    supervise local 3D trajectory prediction, followed by closed-loop tracking
    with a low-level feedback controller.}
    \label{fig:overview}
\end{figure*}

Vision-language-action (VLA) models connect visual understanding and language
to robot behavior~\cite{zitkovich2023rt2,kim2025openvla,qu2025spatialvla},
and recent work has begun to explore this connection in underwater
environments~\cite{wang2025underwatervla,gu2026usimu0,xue2025oceangym}.
Meanwhile, marine vision-language datasets provide rich annotations for
objects, spatial relationships, and scene structure~\cite{xu2025nautilus,
zhu2026usibrain}. These resources offer an opportunity to learn about the
underwater world, but rarely include paired robot trajectories.
Collecting underwater navigation demonstrations remains costly and difficult
to scale, and spatial understanding alone does not specify how a robot
should move. We therefore investigate a central question:
\emph{Can spatial knowledge acquired from large-scale marine observations
be transferred to language-conditioned 3D underwater navigation?}

We propose \textbf{NereidVLA}, an underwater VLA model that connects marine
spatial learning with trajectory-supervised navigation adaptation
(Figure~\ref{fig:overview}). The first stage establishes spatial knowledge
before learning robot motion. Using available annotations and question-answer
formats from NAUTILUS~\cite{xu2025nautilus} and
USI-Brain~\cite{zhu2026usibrain}, together with underwater simulation data,
we construct approximately two million marine spatial-understanding examples.
We use these examples to train \textbf{Sea2Know}, built on
Qwen3-VL-8B~\cite{bai2025qwen3vl}, to learn target grounding, spatial
relationships, and scene structure. This stage makes marine observations
useful for navigation preparation without requiring paired robot-motion
annotations.

The second stage turns this spatially informed model into a trajectory
predictor for closed-loop navigation. Using 573 simulated scenes in
MarineGym~\cite{chu2025marinegym}, we collect 5,117 expert demonstrations
and derive approximately 100,000 navigation examples spanning obstacle
avoidance, target search, pipeline navigation, and cave navigation.
An expert controller executes geometric reference plans under simulated
vehicle dynamics, and the resulting robot motion supplies the trajectory
targets. We use these demonstrations to adapt Sea2Know into NereidVLA,
encoding local body-frame displacements as motion tokens.
Given RGB histories and language instructions, the model autoregressively
predicts local 3D trajectories, which a low-level feedback controller tracks
as visual observations are updated and the trajectory is replanned.

Our evaluation examines both the spatial knowledge learned in the first
stage and its value for navigation. On a six-dimensional underwater
navigation understanding benchmark strictly disjoint from the spatial
training data, Sea2Know outperforms its backbone across all six dimensions.
NereidVLA improves offline trajectory prediction over
SimpleVLN~\cite{hong2026simplevln} in unseen scenes.
Under matched downstream training and execution settings, Sea2Know
initialization increases closed-loop simulation success from 72.4\% to
84.4\% compared with direct navigation fine-tuning of Qwen3-VL-8B.
On a BlueROVHeavy robot, NereidVLA achieves a \textbf{78.75\% task success
rate} in laboratory-pool trials without fine-tuning on the test environment,
compared with 47.50\% for InternVLA-N1.

Our main contributions are as follows:
\begin{enumerate}
    \item We propose NereidVLA, a two-stage underwater VLA framework that
    transfers marine spatial knowledge to language-conditioned local
    3D trajectory prediction and closed-loop navigation.

    \item We construct a diverse underwater navigation dataset from expert
    demonstrations executed under simulated vehicle dynamics, and introduce
    an independent six-dimensional underwater navigation understanding
    benchmark.

    \item We demonstrate improvements in offline trajectory prediction and
    closed-loop navigation in simulation and on a real robot.
    Matched comparisons establish the benefit of marine spatial pretraining,
    while laboratory-pool trials demonstrate transfer to physical execution
    without test-environment fine-tuning.
\end{enumerate}
\section{Related Work}
\label{sec:related}

\noindent\textbf{Underwater vision-language understanding.}
NAUTILUS introduces NautData for underwater captioning, visual
question answering, grounding, and detection, alongside an
imaging-prior-based visual feature enhancement module~\cite{xu2025nautilus}.
USI-Brain develops spatial supervision covering object understanding,
spatial relations, depth reasoning, degradation-aware perception,
and scene understanding~\cite{zhu2026usibrain}.
UWBench evaluates underwater captioning, grounding, and visual
question answering, while MarineEval focuses on marine-domain
knowledge and reasoning~\cite{zhang2025uwbench,wong2026marineeval}.
These efforts advance underwater visual understanding and
evaluation but do not provide paired robot-motion supervision.
We use NAUTILUS and USI-Brain data to train Sea2Know
and introduce an independent benchmark for
navigation-relevant spatial judgments.

\noindent\textbf{Vision-language-action learning and spatial reasoning.}
OpenSpatial generates spatial supervision grounded in
3D object boxes~\cite{liu2026openspatial}, while
SpatialAxiom~\cite{lou2026spatialaxiom} and
SenseNova-SI~\cite{cai2026sensenova,opensensenova2026si13}
advance general-purpose spatial reasoning.
Zhu~\etal improve motion-grounded action understanding through
action-centric video--text alignment~\cite{zhu2026beyondscene}.
RT-2 represents actions as tokens and draws on vision-language
and robotic data~\cite{zitkovich2023rt2},
while OpenVLA provides an adaptable open-source VLA
model~\cite{kim2025openvla}.
SpatialVLA introduces Ego3D encoding and adaptive action grids
for spatially grounded action prediction~\cite{qu2025spatialvla},
and FAST explores efficient action-sequence
tokenization~\cite{pertsch2025fast}.
For navigation, SimpleVLN adapts multimodal models using
visual histories and action chunks~\cite{hong2026simplevln},
while InternVLA-N1 combines high-level goal planning with
low-level execution~\cite{internrobotics2025internvlan1}.
We examine how domain-specific spatial knowledge acquired
before navigation adaptation supports
language-conditioned local 3D trajectory prediction.

\noindent\textbf{Underwater language-conditioned behavior learning.}
UnderwaterVLA combines high-level task reasoning with low-level
control~\cite{wang2025underwatervla}.
USIM provides automatically collected underwater demonstrations,
while U0 incorporates relative-pose supervision for learning
underwater behaviors~\cite{gu2026usimu0}.
AquaCap uses executable control programs and feedback-based
replanning~\cite{li2026aquacap}, while AQUAACT contributes
teleoperation demonstrations and closed-loop policy
evaluation~\cite{wu2026aquaact}.
OceanGym supports language-grounded perception and sequential
decision-making in underwater environments~\cite{xue2025oceangym}.
These approaches explore complementary aspects of underwater
behavior and execution. NereidVLA focuses on transferring
marine spatial knowledge to 3D trajectory prediction through
dynamics-grounded demonstrations, followed by closed-loop navigation.

\noindent\textbf{Underwater navigation and simulation.}
Prior work investigates online 3D path replanning under
motion constraints~\cite{hernandez2019online} and mapless
obstacle avoidance through reinforcement learning~\cite{yang2023collisionfree}.
HoloOcean provides underwater vehicle and sensor
simulation~\cite{potokar2022holoocean}, while OceanSim advances
underwater optical and sonar rendering~\cite{song2025oceansim}.
MarineGym supports GPU-parallel simulation, hydrodynamic
modeling, and domain randomization~\cite{chu2025marinegym}.
Using MarineGym, we construct diverse navigation scenes
and collect expert trajectories under simulated vehicle
dynamics. Geometric paths guide the demonstrations,
but the trajectory supervision comes from executed robot
motion rather than the planned paths themselves.

\section{Method}
\label{sec:method}

\begin{figure*}[!t]
    \centering
    \includegraphics[width=\textwidth]{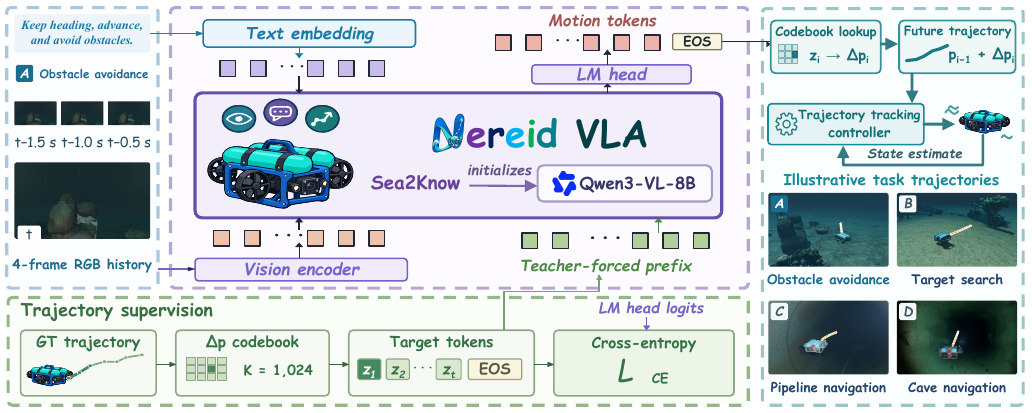}
    \caption{Navigation architecture and trajectory supervision. A
    Sea2Know-initialized navigation VLA predicts discrete motion tokens from
    four historical RGB observations and a task instruction. Executed expert
    trajectories provide token-level supervision, and codebook decoding
    reconstructs local 3D waypoints for closed-loop tracking.}
    \label{fig:navigation-architecture}
\end{figure*}

\subsection{Problem Formulation and Overview}
\label{sec:method-overview}

We formulate language-conditioned underwater navigation as local 3D trajectory
prediction, covering obstacle avoidance, target search, pipeline navigation,
and cave navigation. At decision time $t$, the policy receives four historical
front-facing RGB observations and a language instruction:
\[
    X_t=(I_{t-1.5},I_{t-1.0},I_{t-0.5},I_t,\ell_t).
\]
It predicts a five-second local trajectory,
\[
    \widehat{\mathcal T}_t=\pi_\theta(X_t)
    =\{(k\Delta,\widehat{\mathbf p}_{t,k})\}_{k=1}^{H},
\]
where $H=10$ and $\Delta=0.5\,\mathrm{s}$. All predicted positions are
expressed in the robot body frame fixed at decision time $t$. The VLA uses
only visual observations and language instructions; robot state estimates
are used for trajectory supervision and low-level feedback control,
rather than as policy inputs.

Figure~\ref{fig:overview} summarizes our two-stage framework. We first learn
underwater spatial knowledge in \emph{Sea2Know}, then initialize NereidVLA,
our navigation model, with the learned parameters and adapt it using expert
trajectories. The predicted local
trajectory is decoded into 3D waypoints and tracked by a low-level controller
under closed-loop visual feedback (Figure~\ref{fig:navigation-architecture}).

\subsection{Marine Spatial Knowledge Learning}
\label{sec:marine-spatial-learning}

The first stage learns navigation-relevant spatial knowledge
from marine visual data without paired robot trajectories.
We construct approximately two million vision-language
training examples using resources from
NAUTILUS~\cite{xu2025nautilus} and
USI-Brain~\cite{zhu2026usibrain}, together with underwater
simulation scene data. We reuse available annotations,
including object categories and bounding boxes, and draw
on existing marine question-answer formats to organize
supervision for target grounding, relative spatial
relations, depth ordering, and scene structure.
For sequential observations, we also construct multi-frame
questions about changes in target position, visibility,
and apparent size. Together, these examples provide
supervision for both static spatial relationships and
temporal changes in underwater scenes.

We train Sea2Know through full-parameter supervised
fine-tuning of Qwen3-VL-Instruct backbones at 2B, 4B,
and 8B scales~\cite{bai2025qwen3vl}. This stage learns
visually grounded spatial and temporal representations
without paired robot trajectories or dynamics supervision.
The resulting Sea2Know (8B) provides the initialization
for subsequent trajectory-supervised navigation adaptation.

\subsection{Navigation Demonstration Generation}
\label{sec:method-data}

We build a navigation demonstration pipeline on MarineGym~\cite{chu2025marinegym},
covering five environment families: seabeds, fixed underwater facilities,
pipelines, aquaculture cages, and caves. Each environment varies task goals,
obstacle layouts, and traversable structures, together with water and lighting
conditions.

Expert references are generated according to the geometry of each task.
For obstacle avoidance and target search, we use 3D path search and collision
checks to obtain feasible routes around obstacles or toward targets. For
pipeline and cave navigation, reference paths additionally follow pipe axes
or cave topology while accounting for local obstructions. Candidate routes
are checked against the robot's physical envelope and passage constraints.

We derive trajectory supervision from \emph{executed motion}, rather than
using the planned paths as training targets. A dynamics-based expert
controller tracks each feasible reference with the simulated underwater robot,
while RGB observations,
robot motion, and task information are recorded in temporal alignment. We
retain valid observation--trajectory segments from demonstrations that satisfy
the relevant task and safety checks.

\subsection{Trajectory-Supervised Navigation Adaptation}
\label{sec:method-adaptation}

Motivated by AutoVLA's tokenization of planar
displacements and heading changes~\cite{zhou2025autovla},
we encode underwater motion as 3D body-frame
displacements $(\Delta x,\Delta y,\Delta z)$ to capture
depth changes. We train the Sea2Know-initialized VLA
to predict these motion tokens autoregressively.

\paragraph{Local Trajectory Representation.}
We express all future expert positions in the robot body frame fixed at the
current decision time. Let $\mathbf p_t^W$ be the robot position in world
coordinates and $\mathbf R_t^W$ the body-to-world rotation. The $k$-th target
position is
\begin{equation}
    \mathbf p^*_{t,k}
    =(\mathbf R_t^W)^\top
    \left(\mathbf p^W_{t+k\Delta}-\mathbf p_t^W\right),
    \qquad k=1,\ldots,H.
    \label{eq:local-trajectory-target}
\end{equation}
The body axes follow the forward--left--up convention. Future poses are
used only to construct trajectory targets, not as policy inputs.

\paragraph{Residual-Aware Motion Tokenization.}
We construct a fixed codebook
$\mathcal C=\{\mathbf c_j\}_{j=0}^{K-1}$ with $K=1024$
by clustering consecutive displacements from navigation training
trajectories, expressed in the fixed decision-time body frame.
Rather than quantizing each consecutive displacement independently, we select
a codeword according to the residual between the target and previously
reconstructed positions:
\begin{equation}
\begin{aligned}
    a_{t,k}
    &=\arg\min_{0\le j<K}
      \left\|\mathbf p^*_{t,k}
      -\widetilde{\mathbf p}_{t,k-1}-\mathbf c_j\right\|_2^2,\\
    \widetilde{\mathbf p}_{t,k}
    &=\widetilde{\mathbf p}_{t,k-1}+\mathbf c_{a_{t,k}},
    \qquad \widetilde{\mathbf p}_{t,0}=\mathbf 0.
\end{aligned}
\label{eq:residual-tokenization}
\end{equation}
Here $a_{t,k}$ indexes the selected codeword and
$\widetilde{\mathbf p}_{t,k}$ is the cumulatively reconstructed position.
Conditioning each selection on the current reconstruction allows subsequent
codewords to compensate for earlier quantization errors. We map the indices
to added motion tokens, obtaining
$Y_t^*=(y^*_{t,1},\ldots,y^*_{t,H},\mathrm{EOS})$.
The codebook remains fixed during adaptation and inference.

\paragraph{Autoregressive Trajectory Supervision.}
We initialize the navigation VLA from Sea2Know. A shared visual encoder
processes the RGB history, while the instruction is represented by text
embeddings; both condition the multimodal language model. Its native LM head
predicts motion tokens without a separate action head. With teacher forcing, we optimize
\begin{equation}
    \mathcal L_{\mathrm{traj}}
    =-\frac{1}{H+1}\sum_{k=1}^{H+1}
    \log P_\theta\!\left(y^*_{t,k}\mid X_t,y^*_{t,<k}\right),
    \label{eq:trajectory-loss}
\end{equation}
where $y^*_{t,H+1}=\mathrm{EOS}$. Only the motion tokens and EOS contribute to
the loss; visual and language context positions are masked. Navigation
adaptation uses neither an additional 3D coordinate regression loss nor a
reasoning-text supervision objective.

\paragraph{Trajectory Decoding.}
At inference, predicted motion tokens are decoded
by cumulatively summing their codebook displacements,
producing ten time-indexed local 3D waypoints
for closed-loop execution.

\subsection{Closed-loop Trajectory Execution}
\label{sec:method-execution}

We execute navigation through receding-horizon prediction
and feedback tracking. After following a short segment
of the predicted five-second trajectory, the robot
updates its visual observations and replans.
To account for inference latency, the predicted trajectory
is transformed into a fixed local execution frame using
the estimated robot pose at image capture time.

A low-level controller interpolates the predicted
waypoints and tracks them through state estimation,
PID feedback, and dynamics feedforward.
Since the VLA predicts positions rather than control
commands, the execution module determines attitude
and heading references under vehicle motion constraints
and declared heading rules, and allocates commands to eight thrusters.
In simulation, replanning occurs at 1 Hz of simulated time.

\section{Experiments}
\label{sec:experiments}

We evaluate Sea2Know on underwater spatial understanding and NereidVLA on offline 3D trajectory prediction and closed-loop navigation in simulation and on a real robot. We also assess the contribution of marine spatial pretraining.

\subsection{Experimental Setup}
\label{sec:experimental-setup}

\paragraph{Datasets and evaluation protocols.}
Using MarineGym~\cite{chu2025marinegym} and high-fidelity underwater assets, we construct a training pool of 573 simulated scenes and collect 5,117 expert demonstrations executed under simulated vehicle dynamics. The demonstrations cover obstacle avoidance, target search, pipeline navigation, and cave navigation. We extract 100,000 training samples from 4,597 demonstrations across 516 scenes, with 18,726 samples for validation. Offline prediction uses approximately 10,000 test samples. Both offline testing and closed-loop simulation use 50 independent scenes held out from the navigation training pool, with unseen high-fidelity assets, layouts, and structures.

Our six-dimensional Underwater Navigation Understanding Benchmark covers target grounding (TGA), spatial relations (SRA), passage structure (TSA), traversability prediction (TPA), temporal understanding (TUA), and local action decisions (LADA). TPA and LADA assess visual judgments of traversable space and plausible local actions, without imposing robot-specific motion constraints. Three trained annotators independently label 4,500 candidate questions and cross-check the annotations, retaining 3,646 after filtering disagreements and low-quality items. The benchmark is strictly disjoint from the Stage 1 training data. Annotation details and additional NautData and UWBench~\cite{xu2025nautilus,zhang2025uwbench} results are provided in the supplement.

For closed-loop simulation, we evaluate 25 distinct navigation instances for each of the four tasks. Each instance is tested under five lighting and water conditions, giving 500 executions per model. Models tested on the same robot use the same low-level controller and evaluation conditions.

\paragraph{Baselines and adaptation.}
For spatial understanding, Sea2Know is evaluated after Stage~1;
comparison models are not adapted on our navigation data,
and Qwen3-VL uses its original Instruct checkpoints. For offline
prediction and closed-loop navigation, all baselines use the same
navigation training samples as NereidVLA, except U0 and
$\pi_{0.5}$-USIM, which retain their original underwater
checkpoints for transfer evaluation. For the initialization ablation, Qwen3-VL-8B uses matched
downstream training and execution settings. Model-specific details are provided in the supplement.

\paragraph{Metrics and implementation.}
We report accuracy for spatial understanding; ADE, FDE,
Z MAE, ADE P95, and FDE P95 for trajectory prediction;
and success rate (SR), Grid-SPL (based on
SPL~\cite{anderson2018evaluation}), and Jerk RMS for closed-loop navigation. Grid-SPL measures success-weighted path efficiency against
robot-aware A* reference paths on a 0.25-m 3D grid; Jerk RMS
measures translational motion smoothness. Details are provided
in the supplement.

In Stage~1, Sea2Know models (2B, 4B, and 8B) are initialized from
size-matched Qwen3-VL-Instruct backbones and trained with
full-parameter vision-language supervised fine-tuning, jointly
updating the visual encoder, vision-language merger, and
language model. In Stage~2, NereidVLA is initialized from
Sea2Know (8B) and trained with autoregressive trajectory
supervision. We freeze the visual encoder while fully updating the
language model, token embeddings, and LM head
without LoRA. Further details are given in the supplement.

\begin{figure*}[!t]
    \centering
    \includegraphics[width=\textwidth]{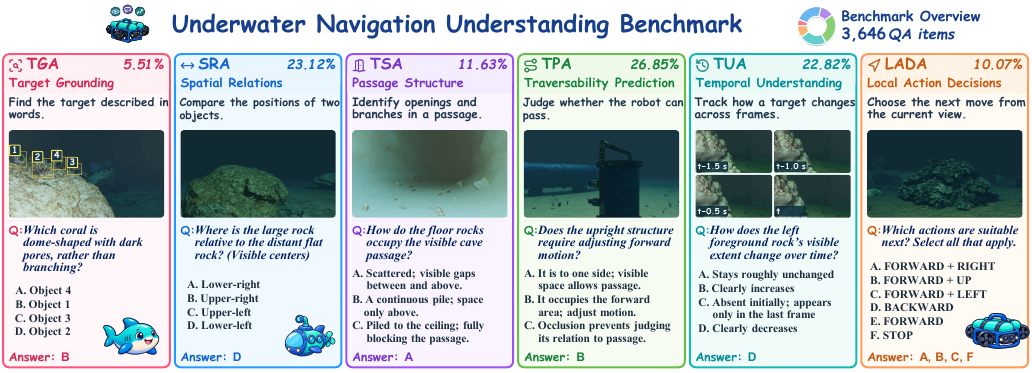}
    \caption{Underwater Navigation Understanding Benchmark. The six panels
    show representative examples and question proportions for
    target grounding, spatial relations, passage structure, traversability,
    temporal understanding, and local action decisions. Each example pairs
    visual observations with a question, answer options, and the reference
    answer.}
    \label{fig:understanding-benchmark}
\end{figure*}

\begin{table*}[!t]
    \centering
    \caption{Task-oriented underwater VQA evaluation (\%). Bold marks the column best; dark/light blue marks the
    best/second-best Sea2Know results. Parentheses show point changes
    from size-matched Qwen3-VL backbones (red/green: gains/drops).}
    \label{tab:task-understanding}
    \begingroup
    \footnotesize
    \setlength{\tabcolsep}{1.8pt}
    \renewcommand{\arraystretch}{1.0}
    \setlength{\aboverulesep}{1pt}
    \setlength{\belowrulesep}{1.3pt}
    \setlength{\fboxsep}{0.6pt}
    \definecolor{SeaBest}{RGB}{203,227,245}
    \definecolor{SeaSecond}{RGB}{234,242,248}
    \definecolor{SeaGain}{RGB}{200,38,38}
    \definecolor{SeaDrop}{RGB}{40,104,74}
    \newcommand{\SeaResult}[3]{%
        \colorbox{#1}{#2\hspace{0.12em}{\scriptsize #3}}}
    \begin{tabular*}{\textwidth}{@{\extracolsep{\fill}}l*{6}{c}@{}}
        \toprule
        {\small Method} & {\small TGA $\uparrow$} & {\small SRA $\uparrow$} & {\small TSA $\uparrow$}
        & {\small TPA $\uparrow$} & {\small TUA $\uparrow$} & {\small LADA $\uparrow$} \\
                \midrule
        \multicolumn{7}{@{}l}{\small \textit{API-based General Models}} \\
        DeepSeek V4.1 Flash~\cite{deepseek2026v41flash} & 87.06 & \textbf{87.66} & \textbf{87.26} & \textbf{86.82} & \textbf{75.48} & \textbf{34.88} \\
        Doubao Seed 2.1 Pro~\cite{bytedance2026seed21} & 87.06 & 84.34 & 84.91 & 84.58 & 72.36 & 32.43 \\
        GPT-6-Sol~\cite{openai2026gpt6sol} & \textbf{91.54} & 82.44 & 80.90 & 77.73 & 69.35 & 33.24 \\
        Claude Opus 5~\cite{anthropic2026opus5} & 85.07 & 82.92 & 84.20 & 83.15 & 71.15 & 26.70 \\
        \midrule
        \multicolumn{7}{@{}l}{\small \textit{Open-source General VLMs}} \\
        Qwen3-VL-32B-Instruct & 63.68 & 72.00 & 72.41 & 78.24 & 57.81 & 28.34 \\
        Qwen3-VL-8B-Instruct & 63.18 & 53.38 & 66.75 & 64.76 & 38.82 & 16.35 \\
        Qwen3-VL-4B-Instruct & 58.21 & 54.21 & 62.97 & 65.88 & 34.98 & 15.80 \\
        Qwen3-VL-2B-Instruct & 38.81 & 42.59 & 48.82 & 60.98 & 36.90 & 16.62 \\
        Qwen3.5-9B~\cite{qwen2026qwen35} & 62.69 & 64.18 & 72.41 & 69.05 & 34.98 & 21.53 \\
        \midrule
        \multicolumn{7}{@{}l}{\small \textit{Spatial-understanding Models}~\cite{cai2026sensenova,opensensenova2026si13}} \\
        SenseNova-SI-1.3-Qwen3-VL-8B & 59.70 & 54.80 & 67.69 & 67.01 & 49.88 & 19.07 \\
        SenseNova-SI-1.3-InternVL3-8B & 50.25 & 33.45 & 59.91 & 59.75 & 35.22 & 14.71 \\
        SpatialAxiom-9B~\cite{lou2026spatialaxiom} & 56.22 & 63.82 & 69.34 & 66.50 & 39.18 & 13.90 \\
        \midrule
        \multicolumn{7}{@{}l}{\small \textit{Underwater-specialized Models}} \\
        OceanGPT-o-7B~\cite{zjunlp2025oceangpto} & 50.75 & 31.79 & 49.53 & 47.19 & 33.65 & 13.35 \\
        NAUTILUS-Qwen-7B & 29.85 & 21.71 & 27.12 & 25.74 & 8.89 & 9.26 \\
        USI-Brain~\cite{zhu2026usibrain} & 51.24 & 31.32 & 53.30 & 50.26 & 35.46 & 8.99 \\
        \midrule
        \multicolumn{7}{@{}l}{\small \textit{Navigation Models}} \\
        SimpleVLN-Qwen3-VL-8B-SFT & 9.45 & 0.00 & 1.18 & 0.51 & 0.12 & 7.90 \\
        \midrule
        \multicolumn{7}{@{}l}{\small \textit{Ours}} \\
        \textbf{Sea2Know (2B)} & \SeaResult{white}{57.21}{\textcolor{SeaGain}{(+18.40)}} & \SeaResult{white}{67.02}{\textcolor{SeaGain}{(+24.43)}} & \SeaResult{white}{72.17}{\textcolor{SeaGain}{(+23.35)}} & \SeaResult{white}{77.02}{\textcolor{SeaGain}{(+16.04)}} & \SeaResult{white}{62.98}{\textcolor{SeaGain}{(+26.08)}} & \SeaResult{white}{19.07}{\textcolor{SeaGain}{(+2.45)}} \\
        \textbf{Sea2Know (4B)} & \SeaResult{SeaSecond}{67.16}{\textcolor{SeaGain}{(+8.95)}} & \SeaResult{SeaSecond}{72.12}{\textcolor{SeaGain}{(+17.91)}} & \SeaResult{SeaSecond}{75.24}{\textcolor{SeaGain}{(+12.27)}} & \SeaResult{SeaSecond}{77.22}{\textcolor{SeaGain}{(+11.34)}} & \SeaResult{SeaSecond}{67.43}{\textcolor{SeaGain}{(+32.45)}} & \SeaResult{SeaSecond}{29.43}{\textcolor{SeaGain}{(+13.63)}} \\
        \textbf{Sea2Know (8B)} & \SeaResult{SeaBest}{67.66}{\textcolor{SeaGain}{(+4.48)}} & \SeaResult{SeaBest}{75.21}{\textcolor{SeaGain}{(+21.83)}} & \SeaResult{SeaBest}{78.30}{\textcolor{SeaGain}{(+11.55)}} & \SeaResult{SeaBest}{77.83}{\textcolor{SeaGain}{(+13.07)}} & \SeaResult{SeaBest}{67.79}{\textcolor{SeaGain}{(+28.97)}} & \SeaResult{SeaBest}{31.61}{\textcolor{SeaGain}{(+15.26)}} \\
        \bottomrule
    \end{tabular*}
    \endgroup
\end{table*}

\begin{table*}[!t]
    \centering
    \caption{Offline trajectory prediction results. All metrics are
    lower-is-better. Boldface marks the best score in each column.}
    \label{tab:trajectory-prediction}
    \begingroup
    \footnotesize
    \setlength{\tabcolsep}{5pt}
    \renewcommand{\arraystretch}{1.15}
    \begin{tabular*}{\textwidth}{@{\extracolsep{\fill}}l*{5}{c}@{}}
        \toprule
        {\small Model} & {\small ADE $\downarrow$} & {\small FDE $\downarrow$} & {\small Z MAE $\downarrow$} & {\small ADE P95 $\downarrow$} & {\small FDE P95 $\downarrow$} \\
        \midrule
        InternVLA-N1 & 0.20008 & 0.39498 & 0.05735 & 0.67412 & 1.26874 \\
        Qwen3-VL-4B & 0.23568 & 0.45514 & 0.06190 & 0.74359 & 1.37255 \\
        Qwen3-VL-2B & 0.28983 & 0.53348 & 0.06178 & 0.91908 & 1.65784 \\
        Qwen3-VL-8B & 0.29121 & 0.55396 & 0.06181 & 0.91134 & 1.66065 \\
        Qwen3-VL-32B & 0.19766 & 0.38996 & 0.05219 & 0.64042 & 1.20181 \\
        NAUTILUS (7B) & 0.21748 & 0.42234 & 0.05230 & 0.74137 & 1.37125 \\
        SimpleVLN (8B) & 0.18501 & 0.37259 & 0.05424
        & \textbf{0.57398} & 1.09097 \\
        \midrule
        \textbf{Ours (8B)} & \textbf{0.18082} & \textbf{0.34970}
        & \textbf{0.04658} & 0.59726 & \textbf{1.06058} \\
        \bottomrule
    \end{tabular*}
    \endgroup
\end{table*}

\begin{table*}[!t]
    \centering
    \caption{Closed-loop navigation in simulation. Each model is evaluated
    on 500 trials (125 per task). Overall SR averages the success rates across the four equally sized tasks.
    Bold marks the best score in each column.
    $\dagger$: original checkpoints without navigation adaptation.}
    \label{tab:simulation-closed-loop}
    \begingroup
    \footnotesize
    \setlength{\tabcolsep}{3pt}
    \renewcommand{\arraystretch}{1.15}
    \begin{tabular*}{\textwidth}{@{\extracolsep{\fill}}l*{4}{c}@{\hspace{14pt}}*{3}{c}@{}}
        \toprule
        \multirow{2}{*}{\small Method}
        & \multicolumn{4}{c}{\small Task success rate (\%) $\uparrow$}
        & \multicolumn{3}{c}{\small Overall performance} \\
        \cmidrule(lr){2-5}\cmidrule(lr){6-8}
        & {\small \shortstack{Obstacle\\avoidance}}
        & {\small \shortstack{Target search\\/ approach}}
        & {\small \shortstack{Pipeline\\navigation}}
        & {\small \shortstack{Cave\\navigation}}
        & {\small SR (\%) $\uparrow$}
        & {\small Jerk RMS $\downarrow$}
        & {\small Grid-SPL (\%) $\uparrow$} \\
        \midrule
        U0$^{\dagger}$ & 5.60 & 19.20 & 10.40 & 13.60 & 12.20 & 4.1906 & 10.71 \\
        $\pi_{0.5}$-USIM$^{\dagger}$~\cite{intelligence2025pi05,gu2026usimu0} & 0.80 & 8.00 & 4.80 & 0.80 & 3.60 & 1.7826 & 3.09 \\
        \midrule
        Qwen3-VL-8B & 84.00 & 52.00 & 71.20 & 82.40 & 72.40 & 1.1247 & 71.52 \\
        Qwen3-VL-32B & 76.00 & 44.00 & 59.20 & 65.60 & 61.20 & 1.4338 & 59.47 \\
        SimpleVLN (8B) & 74.40 & 38.40 & 52.00 & 41.60 & 51.60 & 1.9478 & 49.82 \\
        InternVLA-N1 & 83.20 & 46.40 & \textbf{87.20} & 76.00 & 73.20 & 1.3986 & 70.91 \\
        NAUTILUS (7B) & 79.20 & 47.20 & 83.20 & 61.60 & 67.80 & 1.3621 & 65.62 \\
        \midrule
        \textbf{Ours} & \textbf{85.60} & \textbf{81.60} & 86.40 & \textbf{84.00} & \textbf{84.40} & \textbf{1.0658} & \textbf{81.85} \\
        \bottomrule
    \end{tabular*}
    \endgroup
\end{table*}

\begin{figure*}[!t]
    \centering
    \includegraphics[width=\textwidth]{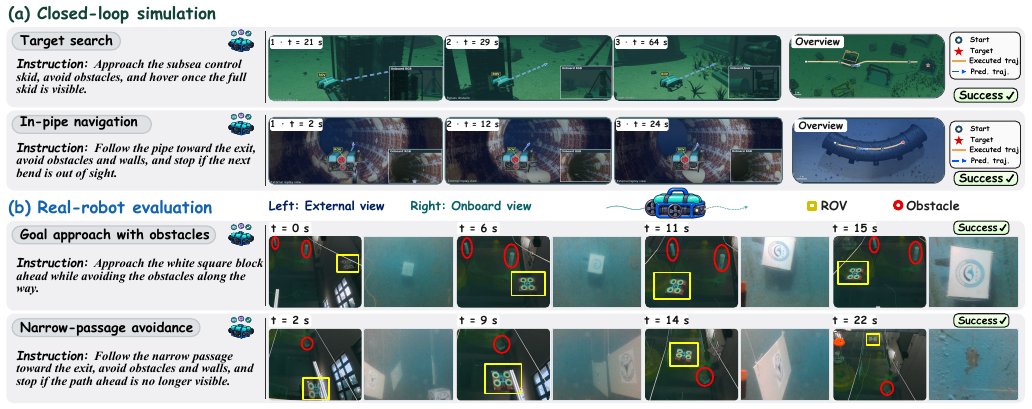}
    \caption{Closed-loop navigation examples. (a) Target search and in-pipe
    navigation in simulation, showing successive frames, onboard views,
    and predicted and executed trajectories. (b) Real-robot goal approach
    with obstacles and narrow-passage avoidance, shown through external
    and onboard views. Time labels indicate stages of each execution.}
    \label{fig:navigation-qualitative}
\end{figure*}

\subsection{Spatial Understanding}
\label{sec:task-understanding}

\paragraph{Navigation-oriented spatial understanding.}
Table~\ref{tab:task-understanding} reports results on our six-dimensional underwater navigation understanding benchmark. Sea2Know (8B) outperforms its Qwen3-VL-8B backbone across all six dimensions, with gains of \textbf{21.83 points in spatial relations, 28.97 in temporal understanding, and 15.26 in local action decisions}. Target grounding improves by a smaller margin of 4.48 points. These gains suggest that marine spatial learning strengthens relational and temporal understanding beyond object recognition and grounding.

The improvements are consistent across model scales, with Sea2Know (2B, 4B, and 8B) outperforming their respective backbones on all six dimensions. Sea2Know (8B) also surpasses SpatialAxiom-9B across all dimensions, particularly in \textbf{temporal understanding (67.79\% vs. 39.18\%) and local action decisions (31.61\% vs. 13.90\%)}. Compared with the navigation-specialized SimpleVLN, Sea2Know (8B) achieves higher scores across all six dimensions on this VQA benchmark, including local action decisions (\textbf{31.61\% vs. 7.90\%}). Among the evaluated open-weight models, Sea2Know (8B) leads on five dimensions; Qwen3-VL-32B performs slightly better on traversability prediction, while several API-based models achieve higher scores.

Figure~\ref{fig:understanding-benchmark} presents representative examples from the benchmark. On the external NautData and UWBench datasets, Sea2Know also performs well on object grounding, spatial relations, and referring localization, although it trails the strongest baseline on some shape- and size-related tasks. Full results are provided in the supplementary material.

\subsection{Offline Trajectory Prediction}
\label{sec:trajectory-prediction}

Table~\ref{tab:trajectory-prediction} presents trajectory prediction results on unseen scenes. NereidVLA achieves the lowest error on four of the five metrics. Compared with SimpleVLN, the closest competitor, it reduces ADE by \textbf{2.3\%}, FDE by \textbf{6.1\%}, and Z MAE by \textbf{14.1\%}.

Lower Z MAE indicates more accurate vertical motion
prediction, important for 3D underwater navigation.

NereidVLA achieves the lowest FDE P95 but a slightly higher ADE P95 than SimpleVLN, indicating lower high-percentile endpoint errors but not uniformly lower trajectory errors.

\subsection{Closed-loop Navigation}
\label{sec:closed-loop-navigation}

\paragraph{Closed-loop simulation.}
We evaluate closed-loop navigation with repeated
observations, replanning, and feedback tracking.
Success requires task completion within the time
limit without safety violations; task-specific
criteria appear in the supplement.

As shown in Table~\ref{tab:simulation-closed-loop},
NereidVLA achieves an overall success rate of
\textbf{84.4\%}, exceeding Qwen3-VL-8B (72.4\%)
by \textbf{12.0 percentage points}. It also achieves
the highest Grid-SPL (81.85\%) and lowest Jerk RMS
(1.0658), indicating favorable path efficiency and
motion smoothness.

Relative to Qwen3-VL-8B, the largest gains occur in target search
(52.0\% to \textbf{81.6\%}) and pipeline navigation
(71.2\% to \textbf{86.4\%}), with smaller gains in obstacle
avoidance and cave navigation. InternVLA-N1~\cite{internrobotics2025internvlan1}
slightly exceeds NereidVLA on pipeline navigation
(87.2\% versus 86.4\%). Figure~\ref{fig:navigation-qualitative}(a)
shows representative predicted and executed trajectories.

\paragraph{Real-robot navigation.}
We deploy NereidVLA on a BlueROVHeavy robot in a laboratory pool without fine-tuning on data from the real test environment. RGB observations are sent to a remote RTX 4090 server for asynchronous planning and local 3D trajectory prediction. The predicted trajectories are then tracked by a low-level controller using IMU-based state feedback.

We evaluate each model on four tasks: target approach, obstacle avoidance, goal approach with obstacles, and narrow-passage avoidance, conducting 20 trials per task with varied target positions and obstacle layouts. A trial is successful if the robot completes the task without collision. The two target-approach tasks require reaching within 1.5 m of the target, while narrow-passage navigation requires crossing the designated endpoint.

As shown in Table~\ref{tab:real-robot-closed-loop}, NereidVLA succeeds in \textbf{63 of 80 trials (78.75\%)}, outperforming InternVLA-N1 (47.50\%) and Qwen3-VL-8B (35.00\%) by \textbf{31.25 and 43.75 percentage points}, respectively.

\begin{table}[!htb]
    \centering
    \caption{Real-robot navigation success rates (\%, $\uparrow$).
    TA: target approach; OA: obstacle avoidance; GA: goal approach with
    obstacles; NP: narrow-passage avoidance. Overall is the four-task mean.}
    \label{tab:real-robot-closed-loop}
    \begingroup
    \footnotesize
    \setlength{\tabcolsep}{2pt}
    \renewcommand{\arraystretch}{1.15}
    \begin{tabular*}{\linewidth}{@{\extracolsep{\fill}}l*{4}{c}@{\hspace{9pt}}c@{}}
        \toprule
        {\small Method} & {\small TA} & {\small OA} & {\small GA} & {\small NP} & {\small Overall} \\
        \midrule
        Qwen3-VL-8B & 35.00 & 80.00 & 15.00 & 10.00 & 35.00 \\
        InternVLA-N1 & 60.00 & 85.00 & 30.00 & 15.00 & 47.50 \\
        \midrule
        \textbf{Ours} & \textbf{90.00} & \textbf{95.00} & \textbf{70.00} & \textbf{60.00} & \textbf{78.75} \\
        \bottomrule
    \end{tabular*}
    \endgroup
\end{table}

NereidVLA achieves \textbf{92.5\%} success across target approach and obstacle avoidance, and \textbf{65.0\%} on the more challenging tasks involving obstacle-aware goal approach and narrow-passage navigation, compared with 22.5\% for InternVLA-N1. Representative executions are shown in Figure~\ref{fig:navigation-qualitative}(b). These results show that NereidVLA can transfer navigation learned from simulated demonstrations to closed-loop execution on a physical underwater robot without adaptation to the test environment.

\subsection{Ablation and Analysis}
\label{sec:ablation-studies}

\paragraph{Effect of marine spatial pretraining.}
To isolate spatial pretraining, we compare
Sea2Know-initialized NereidVLA with Qwen3-VL-8B
fine-tuned directly on the same 3D trajectory
demonstrations, using matched representations,
supervision, and execution settings.
As shown in Table~\ref{tab:trajectory-prediction},
Sea2Know initialization reduces ADE from 0.29121 to
0.18082 (37.9\%) and FDE from 0.55396 to 0.34970
(36.9\%). The gains extend to closed-loop navigation,
where simulation success increases from 72.4\% to 84.4\%
and real-robot success from 35.0\% to 78.75\%.
These results show that marine spatial pretraining
improves 3D trajectory learning beyond what is achieved
through direct navigation fine-tuning alone.

\paragraph{Task-dependent navigation gains.}
The benefits of spatial pretraining vary across tasks. In simulation, success improves by \textbf{29.6 percentage points} in target search and \textbf{15.2 points} in pipeline navigation, compared with 1.6 points in both obstacle avoidance and cave navigation. Target search requires updating the relative
direction of a language-specified goal as views
change, while pipeline navigation relies on
structural direction and nearby free space.
These demands align with Sea2Know's gains in
spatial relations and temporal understanding.

\section{Conclusion}
\label{sec:conclusion}

We presented NereidVLA, a two-stage framework connecting marine
spatial understanding with language-conditioned 3D underwater
navigation. Sea2Know learns navigation-relevant spatial knowledge
from large-scale marine observations; we then adapt it for 3D
motion prediction using dynamics-grounded demonstrations.
Encoding 3D trajectories as discrete motion tokens enables
autoregressive prediction from visual histories and language
instructions, with feedback tracking for closed-loop execution.
Evaluations spanning spatial understanding, offline prediction,
simulation, and real-robot navigation validate this approach.
With matched downstream settings, Sea2Know initialization improves
simulation success from 72.4\% to 84.4\%, while NereidVLA
achieves 78.75\% real-robot success without test-environment
fine-tuning. These results demonstrate that marine spatial
knowledge can be effectively transferred to language-conditioned
3D trajectory prediction and closed-loop underwater navigation.

{\small\raggedright
    \bibliographystyle{ieeenat_fullname}
    \bibliography{references}
}

\ifpreprintappendix
    \clearpage
    \appendix
\setcounter{table}{0}
\renewcommand{\thetable}{A\arabic{table}}
\renewcommand{\theHtable}{appendix.\arabic{table}}

\makeatletter
\setlength{\@fptop}{0pt}
\renewcommand\section{\@startsection{section}{1}{\z@}%
    {-2.5ex plus -0.6ex minus -0.2ex}{1.1ex plus 0.2ex}%
    {\normalfont\large\bfseries\raggedright}}
\makeatother
\section{External Underwater Vision--Language Evaluation}
\label{app:external-vqa}

We evaluate Sea2Know (8B) on NautData~\cite{xu2025nautilus} and
UWBench~\cite{zhang2025uwbench} to examine its spatial understanding
beyond our navigation benchmark. NautData tests grounding and detection;
UWBench covers visual question answering (VQA) and object-referring
localization. Results for Sea2Know, Qwen3.5-9B, and the two SenseNova variants
are from our evaluations; $\dagger$ marks results from the cited work.

On NautData, Sea2Know obtains the highest reported PR@0.5 (55.73) and
detection mAP (19.40), while Qwen3.5-9B has slightly higher grounding
mIoU. On UWBench, Sea2Know leads the position-and-spatial-relations VQA
category (67.24) and all four referring-localization metrics. It trails
the strongest reported model on shape, size, and form questions.

\medskip
\noindent\begin{minipage}{\columnwidth}
    \centering
    \captionof{table}{Grounding and detection on the NautData test split (0--100 scale). $\dagger$: results reported in~\cite{zhu2026usibrain}. Higher is better; best scores are bold.}
    \label{tab:nautdata}
    \begingroup
    \footnotesize
    \setlength{\tabcolsep}{2pt}
    \renewcommand{\arraystretch}{1.15}
    \begin{tabular*}{\columnwidth}{@{\extracolsep{\fill}}l*{4}{c}@{}}
        \toprule
        \multirow{2}{*}{Model}
        & \multicolumn{2}{c}{Grounding}
        & \multicolumn{2}{c}{Detection} \\
        \cmidrule(lr){2-3}\cmidrule(l){4-5}
        & PR@0.5 $\uparrow$ & mIoU $\uparrow$
        & mAP $\uparrow$ & mAP@0.5 $\uparrow$ \\
        \midrule
        \shortstack[l]{SenseNova-SI-1.3-\\Qwen3-VL-8B} & 54.97 & 50.47 & 13.57 & 23.09 \\
        \shortstack[l]{SenseNova-SI-1.3-\\InternVL3-8B} & 5.70 & 8.57 & 0.47 & 0.83 \\
        Qwen3.5-9B & 55.63 & \textbf{51.32} & 18.97 & 30.23 \\
        Qwen3-VL-8B$^\dagger$ & 45.20 & 43.90 & 14.80 & 24.00 \\
        \midrule
        \textbf{Sea2Know (8B)} & \textbf{55.73} & 50.39 & \textbf{19.40} & \textbf{35.20} \\
        \bottomrule
    \end{tabular*}
    \endgroup
\end{minipage}\par

\medskip
\noindent\begin{minipage}{\columnwidth}
    \centering
    \captionof{table}{VQA accuracy (\%) on two UWBench categories. $\dagger$: published results from~\cite{zhang2025uwbench}. Best scores are bold.}
    \label{tab:uwbench-vqa}
    \begingroup
    \footnotesize
    \setlength{\tabcolsep}{3pt}
    \renewcommand{\arraystretch}{1.15}
    \begin{tabular*}{\columnwidth}{@{\extracolsep{\fill}}l*{2}{c}@{}}
        \toprule
        Model
        & \shortstack{Pos. \&\\Spatial Rel.}
        & \shortstack{Shape, Size\\\& Form} \\
        \midrule
        Gemini-2.5-Flash$^\dagger$ & 61.55 & 71.42 \\
        Qwen2.5-VL-7B$^\dagger$ & 55.66 & 64.42 \\
        GLM-4.1V-9B$^\dagger$ & 61.27 & 71.98 \\
        GLM-4.5V-106B$^\dagger$ & 62.48 & \textbf{72.52} \\
        \shortstack[l]{SenseNova-SI-1.3-\\Qwen3-VL-8B} & 55.89 & 64.87 \\
        \shortstack[l]{SenseNova-SI-1.3-\\InternVL3-8B} & 39.67 & 56.94 \\
        \midrule
        \textbf{Sea2Know (8B)} & \textbf{67.24} & 66.96 \\
        \bottomrule
    \end{tabular*}
    \endgroup
\end{minipage}\par

\medskip
\noindent\begin{minipage}{\columnwidth}
    \centering
    \captionof{table}{Object-referring localization on UWBench (0--100 scale). $\dagger$: published results from~\cite{zhang2025uwbench}. Higher is better; best scores are bold.}
    \label{tab:uwbench-referring}
    \begingroup
    \footnotesize
    \setlength{\tabcolsep}{2pt}
    \renewcommand{\arraystretch}{1.15}
    \begin{tabular*}{\columnwidth}{@{\extracolsep{\fill}}l*{4}{c}@{}}
        \toprule
        Model
        & \shortstack{Acc@\\IoU$_{0.5}$ $\uparrow$}
        & \shortstack{Acc@\\IoU$_{0.7}$ $\uparrow$}
        & mIoU $\uparrow$
        & \shortstack{Cum\\IoU $\uparrow$} \\
        \midrule
        Qwen2.5-VL-7B$^\dagger$ & 37.15 & 11.76 & 32.77 & 33.69 \\
        Qwen2.5-VL-72B$^\dagger$ & 90.90 & 74.71 & 77.90 & 81.50 \\
        Qwen3.5-9B & 95.68 & 92.69 & 89.15 & 90.53 \\
        GLM-4.5V-106B$^\dagger$ & 89.21 & 75.38 & 79.26 & 80.51 \\
        \shortstack[l]{InternVL3.5-\\241B-A28B}$^\dagger$ & 84.76 & 64.14 & 71.99 & 76.98 \\
        \shortstack[l]{SenseNova-SI-1.3-\\Qwen3-VL-8B} & 71.16 & 66.65 & 64.72 & 61.22 \\
        \shortstack[l]{SenseNova-SI-1.3-\\InternVL3-8B} & 24.45 & 20.77 & 21.77 & 23.70 \\
        \midrule
        \textbf{Sea2Know (8B)} & \textbf{96.58} & \textbf{93.43} & \textbf{89.95} & \textbf{91.35} \\
        \bottomrule
    \end{tabular*}
    \endgroup
\end{minipage}\par
\section{Data Construction and Understanding Benchmark}
\label{app:data-and-benchmark}

\subsection{Marine spatial training data}
\label{app:marine-spatial-data}
Sea2Know is trained on approximately two million marine vision--language
examples from NAUTILUS~\cite{xu2025nautilus}, USI-Brain~\cite{zhu2026usibrain},
and underwater simulation scenes. Existing annotations and question--answer
formats provide supervision for target grounding, spatial relations, depth
ordering, and scene structure. Multi-frame questions additionally describe
changes in target position, visibility, and apparent size. None of these
examples requires paired robot trajectories.

The Underwater Navigation Understanding Benchmark is disjoint from
Stage~1 training. The external NautData test split is also excluded.

\subsection{Understanding benchmark}
\label{app:understanding-protocol}
Our benchmark evaluates six navigation-related judgments
(Figure~\ref{fig:understanding-benchmark}): target grounding (TGA),
spatial relations (SRA), passage structure (TSA), traversability
prediction (TPA), temporal understanding (TUA), and local action
decisions (LADA). TUA uses four-frame observations. TPA and LADA assess
visually plausible passage and action choices, rather than vehicle
motion under dynamics constraints.

Three trained annotators independently labeled 4,500 candidate
questions and cross-checked the annotations. After removing disputed and low-quality items,
3,646 remain. Accuracy follows the answer format of each question. For
multi-answer LADA questions, a prediction is correct only if its
selected options exactly match the reference set.

\subsection{Navigation demonstrations}
\label{app:navigation-scenes}
We construct navigation scenes in MarineGym~\cite{chu2025marinegym}
across five environments: seabeds, fixed underwater facilities, pipelines,
aquaculture cages, and caves. The training pool contains 573 scenes and
5,117 expert demonstrations. From 4,597 demonstrations in 516 scenes,
we extract 100,000 training samples, with 18,726 additional validation
samples. Offline evaluation uses approximately 10,000 test samples;
both offline and closed-loop tests use 50 independent held-out scenes
with unseen assets, layouts, and structures.

Geometric search and collision checking produce task-specific reference
paths. A controller follows each path under simulated vehicle dynamics
while RGB images, instructions, and vehicle motion are recorded in
temporal alignment. We retain valid segments and use the \emph{executed}
trajectories, not the planned paths, as training targets.
\section{Training and Baseline Adaptation}
\label{app:training-baselines}

\subsection{Stage 1: Sea2Know training}
\label{app:stage-one-training}
We train Sea2Know at 2B, 4B, and 8B scales through full-parameter
supervised fine-tuning of size-matched Qwen3-VL-Instruct backbones.
The vision encoder, vision--language merger, and language model are all
updated. Sea2Know (8B) initializes the navigation model.

\subsection{Stage 2: navigation adaptation}
\label{app:stage-two-training}
Given four RGB observations and a language instruction, NereidVLA
predicts ten motion tokens followed by EOS, representing a five-second
local 3D trajectory. We freeze the vision encoder and fully update the
language model, token embeddings, and LM head, without LoRA. The loss is
cross-entropy over the motion tokens and EOS; visual and text context
tokens are masked. We use neither coordinate-regression nor reasoning-text
supervision.

Training uses 100,000 samples for one epoch (391 optimizer steps) on
32 NVIDIA A800 80GB GPUs across four nodes. A microbatch of one per GPU
and eight gradient-accumulation steps give a global batch size of 256.
We use AdamW ($\beta_1=0.9$, $\beta_2=0.999$; weight decay $0.01$) with
a learning rate of $10^{-5}$, linearly warmed up over 80 steps and then
held constant. Training uses BF16 and fully sharded data parallelism
(FSDP). A fixed subset of 1,024 validation samples is used for monitoring.

\subsection{Motion tokenization}
\label{app:token-implementation}
The ten future positions are expressed in the forward--left--up body
frame fixed at decision time. We cluster displacements from the navigation
training set into a fixed codebook of 1,024 entries. At each step, the
token is chosen from the residual between the target position and the
previously reconstructed waypoint, allowing subsequent tokens to correct
quantization errors. At inference, cumulative codebook displacements
recover the local 3D trajectory. The encoding equations are given in the
main Method section.

\subsection{Baseline adaptation}
\label{app:baseline-implementation}
For offline trajectory prediction and closed-loop navigation, Qwen3-VL
(2B, 4B, 8B, and 32B), SimpleVLN, InternVLA-N1, and NAUTILUS are
adapted using the same navigation training samples as NereidVLA.
U0 and $\pi_{0.5}$-USIM are evaluated in closed-loop simulation from
their original underwater checkpoints, without additional navigation
fine-tuning. These comparisons share training data where applicable,
but not necessarily model architectures or optimization settings.

To isolate spatial initialization, we compare NereidVLA with
Qwen3-VL-8B trained directly on the same trajectory data under matched
trajectory representation, supervision, and execution settings. Only
the initialization differs. For the spatial-understanding benchmark,
comparison models are not adapted on our navigation data; Qwen3-VL
uses its original Instruct checkpoints.
\section{Evaluation Protocols and Metrics}
\label{app:evaluation-protocols}

\subsection{Offline trajectory prediction}
\label{app:offline-metrics}
We compare predicted and expert trajectories in the body frame fixed
at decision time. For $H=10$ waypoints, the errors are
\begin{equation}
\begin{aligned}
    \mathrm{ADE}&=\frac{1}{H}\sum_{k=1}^{H}
      \|\widehat{\mathbf p}_{k}-\mathbf p^*_{k}\|_2,\\
    \mathrm{FDE}&=\|\widehat{\mathbf p}_{H}-\mathbf p^*_{H}\|_2,\\
    \mathrm{Z\ MAE}&=\frac{1}{H}\sum_{k=1}^{H}
      |\widehat p_{k,z}-p^*_{k,z}|.
\end{aligned}
\label{eq:app-offline-metrics}
\end{equation}
ADE P95 and FDE P95 are the 95th percentiles of the respective
per-sample errors across the test set.

\subsection{Closed-loop simulation protocol}
\label{app:simulation-tasks}
We evaluate 25 distinct instances of each of the four navigation tasks
under five water and lighting conditions, totaling 500 runs per model.
The model replans at 1\,Hz of simulated time. A run succeeds if the
vehicle reaches the task goal within the time limit and without violating
safety constraints. Collisions and timeouts count as failures. Success
is determined from executed motion, not predicted waypoints.

\subsection{Grid-SPL}
\label{app:metric-grid-spl}
Grid-SPL weights navigation success by path efficiency relative to a
robot-aware A* reference on a 3D occupancy grid:
\begin{equation}
    \mathrm{Grid\mbox{-}SPL} = \frac{100}{N}\sum_{i=1}^{N}
      S_i\frac{L_i^{\mathrm{grid}}}{\max(L_i,L_i^{\mathrm{grid}})}.
    \label{eq:app-grid-spl}
\end{equation}
Here $S_i\in\{0,1\}$ denotes success, $L_i$ is the executed path
length measured from 100\,Hz simulated positions, and
$L_i^{\mathrm{grid}}$ is the A* reference length ($N=500$).
Reference paths use a 0.25\,m, 26-connected grid, each task's valid goal
region, and collision checks against a sphere enclosing the vehicle.
Dynamic fish affect execution safety but are excluded from the static
reference. Failed runs contribute zero. The grid reference approximates,
rather than exactly solves, continuous-space shortest-path planning.

\subsection{Translational jerk RMS}
\label{app:metric-jerk}
We measure motion smoothness from the jerk of the executed world-frame
translation. For run $i$, jerk is the second derivative of linear
velocity, and the reported score is the mean of per-run RMS values:
\begin{equation}
\begin{aligned}
    \mathbf j_{i,k} &={}
      \left.\frac{\mathrm d^2\mathbf v_i(t)}{\mathrm dt^2}\right|_{t=t_k},\\
    J_i&=\sqrt{\frac{1}{M_i}\sum_{k=1}^{M_i}
      \|\mathbf j_{i,k}\|_2^2},\\
    \overline J_{\mathrm{RMS}}&=\frac{1}{N}\sum_{i=1}^{N}J_i.
\end{aligned}
\label{eq:app-jerk}
\end{equation}
We estimate derivatives of 50\,Hz world-frame velocity using an
11-sample, third-order Savitzky--Golay fit (0.2\,s span).
Each RMS includes all three translational axes. The reported value
(m/s$^3$) averages all runs, including failures. Derivatives use
simulated time, excluding inference-related wall-clock pauses.
\section{Physical Robot System and Evaluation}
\label{app:real-robot-protocol}

\subsection{Hardware and execution}
\label{app:real-robot-hardware}
We deploy a BlueROVHeavy with eight thrusters in a laboratory pool.
RGB observations from a RealSense D435i are processed asynchronously
on a remote RTX 4090 server. The predicted waypoints are tracked by a
low-level controller using IMU-supported state estimation, PID feedback,
and dynamics feedforward.

To account for inference latency, each trajectory is transformed into
a fixed execution frame using the estimated robot pose at image-capture
time. The robot follows a short segment before acquiring new observations
and replanning. No data from the real test environment are used for
fine-tuning.

\subsection{Tasks and success criteria}
\label{app:real-robot-tasks}
We test four tasks with 20 trials per task and model, varying target
positions and obstacle layouts. Success requires completing the task
without collision. Target approach (TA) and goal approach with
obstacles (GA) require reaching within 1.5\,m of the target.
Narrow-passage avoidance (NP) requires crossing the specified endpoint;
obstacle avoidance (OA) requires reaching its assigned objective.
All models share the low-level controller and evaluation conditions.
Results are reported in Table~\ref{tab:real-robot-closed-loop}.
\section{API Model Identifiers}
\label{app:api-model-identifiers}

Table~\ref{tab:api-model-identifiers} lists the requested identifiers
for the API-based spatial-understanding baselines in
Table~\ref{tab:task-understanding}.

\medskip
\noindent\begin{minipage}{\columnwidth}
    \centering
    \captionof{table}{API model identifiers used for the spatial-understanding baselines.}
    \label{tab:api-model-identifiers}
    \begingroup
    \footnotesize
    \setlength{\tabcolsep}{2pt}
    \renewcommand{\arraystretch}{1.15}
    \begin{tabular*}{\linewidth}{@{\extracolsep{\fill}}ll@{}}
        \toprule
        {\small Model} & {\small Requested model ID} \\
        \midrule
        Doubao Seed 2.1 Pro & \textrm{doubao-seed-2-1-pro-260915} \\
        DeepSeek V4.1 Flash & \textrm{deepseek-v4.1-flash-1017} \\
        GPT-6-Sol & \textrm{gpt-6-sol} \\
        Claude Opus 5 & \textrm{claude-opus-5} \\
        \bottomrule
    \end{tabular*}
    \endgroup
\end{minipage}\par

\fi

\end{document}